\documentclass{article}

\usepackage{PRIMEarxiv}

\usepackage[utf8]{inputenc} 
\usepackage[T1]{fontenc}    
\usepackage{hyperref}       
\usepackage{url}            
\usepackage{booktabs}       
\usepackage{amsfonts}       
\usepackage{nicefrac}       
\usepackage{microtype}      
\usepackage{lipsum}
\usepackage{fancyhdr}       
\usepackage{graphicx}       
\graphicspath{{media/}}     

\usepackage{amsmath,amsfonts}
\usepackage{algorithmic}
\usepackage{graphicx}
\usepackage{graphics}
\usepackage{multirow}
\usepackage[ruled,vlined,linesnumbered,algo2e,noend]{algorithm2e}
\DeclareMathOperator{\KL}{KL}
\DeclareMathOperator{\simm}{sim}
\DeclareMathOperator{\softmax}{softmax}
\DeclareMathOperator{\transformer}{Transformer\_Encoder}

\title{Domain Invariant Representation Learning and Sleep Dynamics Modeling for Automatic Sleep Staging}

\author{
  Seungyeon Lee\thanks{Both authors contributed equally to this research}, Thai-Hoang Pham\footnotemark[1] \\
  The Ohio State University \\
  \texttt{\{lee.10029,pham.375\}@osu.edu} \\
   \And
  Zhao Cheng \\
  Google \\
  \texttt{leochengzhao@gmail.com} \\
  \And
  Ping Zhang\thanks{Corresponding author} \\
  The Ohio State University \\
  \texttt{zhang.10631@osu.edu}
}

\begin{document}
\maketitle

\begin{abstract}
Sleep staging has become a critical task in diagnosing and treating sleep disorders to prevent sleep-related diseases. With growing large-scale sleep databases, significant progress has been made toward automatic sleep staging. However, previous studies face critical problems in sleep studies; the heterogeneity of subjects' physiological signals, the inability to extract meaningful information from unlabeled data to improve predictive performances, the difficulty in modeling correlations between sleep stages, and the lack of an effective mechanism to quantify predictive uncertainty. In this study, we propose a neural network-based sleep staging model, DREAM, to learn domain generalized representations from physiological signals and models sleep dynamics. DREAM learns sleep-related and subject-invariant representations from diverse subjects’ sleep signals and models sleep dynamics by capturing interactions between sequential signal segments and between sleep stages. We conducted a comprehensive empirical study to demonstrate the superiority of DREAM, including sleep stage prediction experiments, a case study, the usage of unlabeled data, and uncertainty. Notably, the case study validates DREAM's ability to learn generalized decision function for new subjects, especially in case there are differences between testing and training subjects. Uncertainty quantification shows that DREAM provides prediction uncertainty, making the model reliable and helping sleep experts in real-world applications.
\end{abstract}


\section{Introduction}
Sleep disorders, which are prevalent in the general population, have a negative impact on human mental and physical health, and can further cause serious health problems such as brain stroke, high blood pressure, or myocardial infarction~\cite{zarei2018automatic,luyster2012sleep}. Therefore, sleep studies play an important role in evaluating and improving the quality of sleep, and in particular, the identification of sleep stages has become a critical task for diagnosing and treating sleep disorders. In general, sleep disorders are diagnosed via polysomnography (PSG), which consists of multiple physiology signals including electroencephalography (EEG), electrooculography (EOG), electromyography (EMG),  and electrocardiogram (ECG)~\cite{memar2017novel}. Sleep experts identify sleep stages at 30-second intervals (i.e., sleep segments) in PSG (Figure 1) and diagnose the sleep disorders. However, manually annotating these sleep segments is time-consuming, labor-intensive, costly, and prone to human error.

\begin{figure}[t]
    \centering
    \includegraphics[width=.5\linewidth]{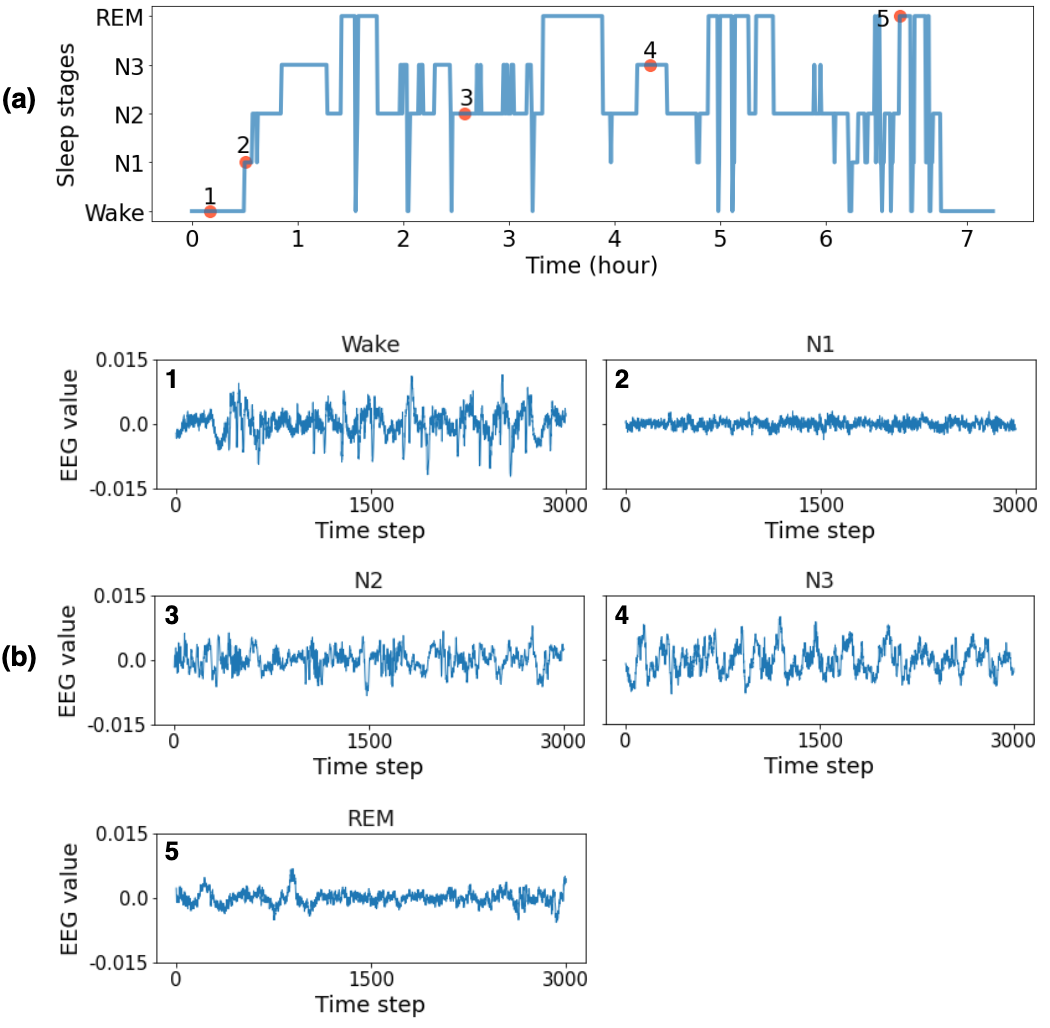}
    \caption{EEG signal visualization. (a) A sleep cycle of a subject in the SleepEDF-20 dataset. Each point in the cycle represents a 30s-interval sleep segment. (b) Details of EEG signal for different sleep segments (i.e., the points in (a)). Each segment is sampled at 100 Hz so the total number of time steps for each segment is 3000.}
    \label{fig:1}
\end{figure}

With rapidly growing large-scale public sleep databases~\cite{o2014montreal,zhang2018national} and advancements in machine learning, significant progress has been made towards automatic sleep staging that automatically predicts sleep stages from given signal datasets. Recently, many deep learning models have been proposed for automatic sleep staging and have shown their superiority over conventional machine learning models. Most models leverage recurrent neural networks (RNNs)~\cite{phan2018automatic,phan2019seqsleepnet,guillot2020dreem} and convolutional neural networks (CNNs) ~\cite{fernandez2020convolutional,dut2021automatic,chambon2018deep, eldele2021attention,perslev2019u}. Additionally, model architectures composed of both CNNs and RNNs are also widely used~\cite{phan2021xsleepnet,mousavi2019sleepeegnet,supratak2020tinysleepnet, supratak2017deepsleepnet}, where CNNs are usually used to learn useful representations for physiological signals, and RNNs are utilized to capture the temporal relationship between them. Unlike conventional machine learning models, deep learning models can automatically extract and learn meaningful features without domain knowledge, making them more feasible in practical settings. However, while existing methods have shown promising results for automatic sleep staging, they have not sufficiently addressed several critical challenges in sleep studies, thereby degrading their predictive performances and hindering downstream applications. These challenges include: 

\begin{itemize}
    \item C1. Physiological signals are significantly different between diverse subjects, depending on the subject’s information such as underlying health conditions. Therefore, each subject can be defined as one domain where its signals follow a distinct distribution. Learning sleep-related but subject-invariant representations for physiological signals is critical for generalized and accurate sleep staging to new subjects.
    \item C2. Most existing methods can only leverage labeled signal datasets, which are limited and costly to train machine learning models. By contrast, massive unlabeled signals are easy to collect in clinical settings. Exploiting meaningful information from unlabeled data can help to boost sleep staging prediction performance.
    \item C3. Sleep stages have a strong transition~\cite{kim2009markov}. Modeling complex interactions between sleep stages at both representation learning and sleep stage prediction levels is essential for accurate sleep staging.
    \item C4. Human intervention is required when applying machine learning in the clinical context. The desired machine learning systems need to provide not only predictions but also the corresponding uncertainty scores to allow clinical experts to intervene when needed.
\end{itemize}

In this study, we propose a novel deep learning-based framework for automatic sleep staging to address these challenges in sleep studies. The proposed method named Domain invaRiant and contrastivE representAtion for sleep dynaMics (DREAM) is composed of a feature representation network and a sleep stage classification. Specifically, we design the feature representation network under domain generalization to learn generalized and robust representations for various subjects based on a variational auto-encoder (VAE) architecture and contrastive learning. The VAE-based model, whose encoder and decoder are constructed from a residual neural network (ResNet)~\cite{he2016deep}, learns the disentangled representation in which the sleep- and subject-related features are separately captured in given sleep signal segments (address C1). To further strengthen the robustness of the representations, we also incorporate contrastive learning into the VAE-based model, and the contrastive learning is applied on both labeled and unlabeled datasets (address C2). We first optimize the feature network and then use the trained feature network with fixed weights to train the sleep stage classification network composed of Transformer~\cite{vaswani2017attention} and conditional random field (CRF)~\cite{lafferty2001conditional}. Given a sequence of sleep signal segments, the trained feature network is utilized to extract a sequence of subject-invariant and sleep-related representations. The extracted sequence is fed into the Transformer model that captures the interactions between these sequential representations and learns the contextualized representation (address C3). Then, the CRF model uses the representation from the Transformer as input to find the sequence of best corresponding sleep stages, instead of finding labels for sleep segments independently (address C3). Finally, the transition of sleep stages learned by the CRF model is used to quantify the uncertainty of the predictions (address C4).
In summary, our contributions follow as:

\begin{itemize}
    \item We propose a novel deep learning-based automatic sleep staging framework named DREAM that learns domain generalized representations from various subjects’ signal segments and models sleep dynamics by capturing complex interactions between sequential sleep segments and between sleep stages.
    \item We develop a feature representation network to learn sleep-related and subject-invariant representation from various subjects in both labeled and unlabeled datasets for generalized predictive performance on new subjects.
    \item We design a sleep stage classification network to explicitly model interactions between sequential signal segments and correlations between sleep stages at the level of representation learning and sleep stage prediction, respectively.
    \item We construct a mechanism that quantifies the uncertainty measures for sleep stage predictions, thereby allowing sleep specialists to effectively apply DREAM in practice.
    \item We conduct a comprehensive empirical study to demonstrate the advantages of the proposed DREAM.

\end{itemize}

Note that DREAM was introduced in our previous conference paper~\cite{lee2022dream}. The key differences between this paper and the prior conference paper are as follows:
\begin{itemize}
    \item We propose a new mechanism that measures the uncertainty of predictions to demonstrate reliability in real-world scenarios.
    \item We introduce an approach that leverages unlabeled EEG data into DREAM to improve predictive performance.
    \item We conduct more comprehensive experiments with a case study, uncertainty quantification, the usage of unlabeled data, an ablation study, and transferability to evaluate the effectiveness of DREAM.
    \item Code is available at a code repository\footnote{\url{https://github.com/yeon-lab/DREAM}}.
\end{itemize}

\section{Related works}
This section briefly reviews existing work relevant to our study, including automatic sleep staging, robustness under data shift, and contrastive learning.

\subsection{Automatic sleep staging}
With rapidly growing large-scale public sleep databases~\cite{o2014montreal,zhang2018national} and advancements in machine learning, significant progress has been made towards automatic sleep staging that automatically predicts sleep stages from given signal datasets. Early attempts leverage conventional machine learning models such as Naive Bayes~\cite{dimitriadis2018novel}, support vector machine~\cite{zhu2014analysis,seifpour2018new}, and random forest~\cite{memar2017novel,li2017hyclasss}. The disadvantage of these models is that they require domain knowledge to manually select the good features used in their predictions from time and frequency spaces. Recently, many deep learning models have been proposed for automatic sleep staging and have shown their superiority over conventional machine learning models. Most models are based on recurrent neural networks (RNNs)~\cite{phan2018automatic,phan2019seqsleepnet,guillot2020dreem} and convolutional neural networks (CNNs) ~\cite{fernandez2020convolutional,dut2021automatic,chambon2018deep, eldele2021attention,perslev2019u}. For example, U-time~\cite{perslev2019u} leverages a fully convolutional autoencoder with skip connections to learn temporal representation. U-time maps input sequences of varying lengths into sequences of class labels, allowing flexibility in choosing the temporal scale. AttnSleep~\cite{eldele2021attention} consists of a feature extraction module and a temporal context encoder. The extraction module combines a multi-resolution CNN to extract low and high-frequency features and adaptive feature recalibration to model interdependencies. The encoder utilizes multi-head attention with causal convolutions to capture temporal dependencies among features. Additionally, model architectures composed of both CNNs and RNNs are also widely used~\cite{phan2021xsleepnet,mousavi2019sleepeegnet,supratak2020tinysleepnet, supratak2017deepsleepnet}, where CNNs are usually used to learn useful representations for physiological signals, and RNNs are utilized to capture the temporal relationship between them. DeepSleepNet~\cite{supratak2017deepsleepnet} utilizes two branches of CNN layers with different sizes for representation learning and LSTM layers for classification. Among the two CNN branches, the smaller and larger CNNs are specialized in extracting temporal characteristics and frequency information, respectively. Unlike conventional machine learning models, deep learning models can automatically extract and learn meaningful features without domain knowledge, making them more feasible in practical settings. However, while existing methods have shown promising results for automatic sleep staging, they have not sufficiently addressed several critical challenges in sleep studies, thereby degrading their predictive performances and hindering downstream applications.
These existing models cannot (i) explicitly model subject- and sleep-specific features to help them can be generalized well in the new subjects, (ii) effectively utilize easy-to-collect unlabeled sleep signal data to improve prediction performances, and (iii) capture the transition between sleep stages at both feature representation and label classification levels.

\subsection{Robustness under data shift} 
Most machine learning models depend on the strong assumption that data points data follows an identical and independent distribution in both training and testing environments. However, real-world scenarios often violate this over-simplified assumption, causing challenges like the out-of-distribution problem. For instance, data distribution of train and test subjects may be different due to their different underlying health conditions. Consequently, models experience significant performance drops during testing. To address this, various methods aim to guarantee the model robustness under the change of environment in different scenarios including (i) transfer learning in which models have full access to both training and testing environment~\cite{weiss2016survey,yosinski2014transferable,donahue2014decaf}, (ii) domain adaptation in which models can access to unlabeled data of testing environment~\cite{lu2020stochastic,saito2018maximum,ganin2015unsupervised}, (iii) domain generalization in which no information about the testing environment is available during training~\cite{muandet2013domain,li2018domain,li2018deep}. In our study, we design the model for the domain generalization scenario so that it can successfully generalize its sleep stage predictions to the new testing subjects based on information from subjects in the training set.

\subsection{Contrastive learning} 
Labeled data which requires many human efforts to manually label is expensive and hard to be scaled up. Recently, many methods have been proposed to utilize unlabeled data which is generally easy to collect to improve the prediction performances and contrastive learning is one of the most powerful approaches among them. The goal of contrastive learning is to construct a representation space in which similar sample pairs stay close to each other while dissimilar ones are far apart and this method can be applied for both supervised~\cite{chen2020simple, van2018representation, zhu2020deep, qian2021spatiotemporal} and unsupervised~\cite{khosla2020supervised} settings. In our study, we leverage the contrastive objective function to learn more robust representations for sleep signal segments from both labeled and unlabeled datasets.

\begin{figure}[t]
    \centering
    \includegraphics[width=1\linewidth]{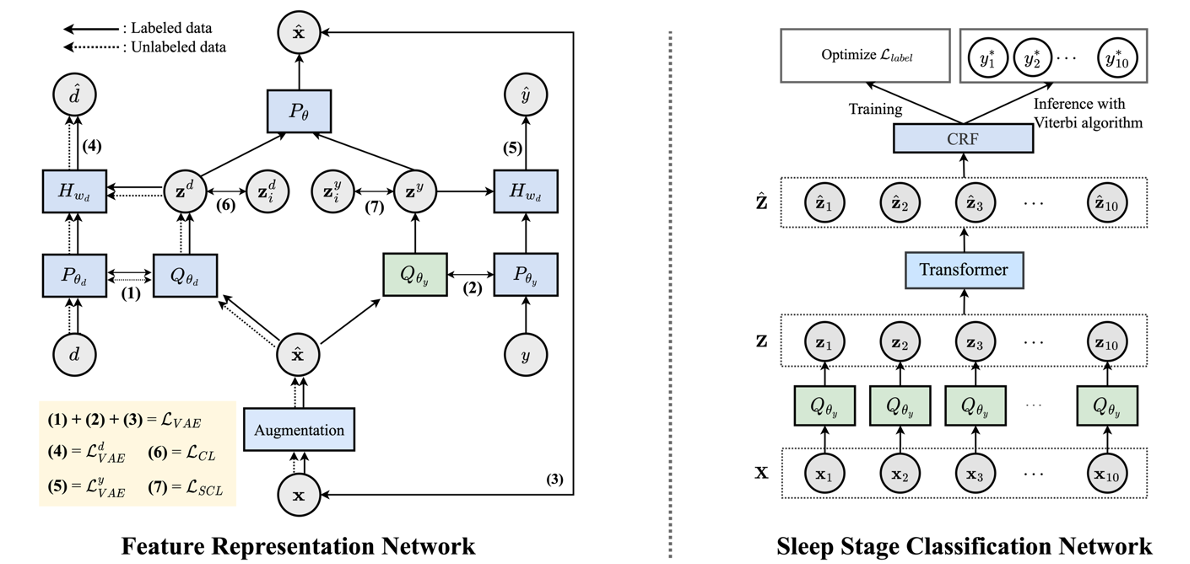}
    \caption{The overall architecture of DREAM. DREAM is composed of a feature representation network and a sleep stage classification network.}
    \label{fig:2}
\end{figure}

\section{Methods}
In this section, upper-case and bold letters (e.g., $\textbf{X}$), lower-case and bold letters (e.g., $\textbf{x}$), and lower-case letters (e.g., $x$) represent matrices, vectors, and scalar, respectively. 

\subsection{Definitions and basic notations for automatic sleep staging}
\begin{itemize}
    \item Sleep signal sequence: The sleep signal sequence from EEG is divided into 30-second segments. $i$-th segment of the $k$th sequence is expressed as $\textbf{x}_{k,i} \in R^N$, where $N=30 \times F$ and $F$ is a sampling rate per second. Then, the $k$th sleep signal sequence is represented as $\textbf{X}_k=[ \textbf{x}_{k,1}, \textbf{x}_{k,2},\cdots , \mathbf{x}_{k,T}]$, where $T$ is the number of segments in the $k$th sequence.
    \item Sleep stage: The sleep stages are classified into three main groups according to the American Academy of Sleep Medicine (AASM) standard: wakefulness (W), rapid eye movement (REM), and non-rapid eye movement (NREM). NREM can be further classified as N1, N2, N3, and N4. According to previous studies, N4 is merged into N3 due to the rarity of N4. The final sleep stage set $\mathcal{Y}$ consists of W, REM, N1, N2, and N3 stages. 
    \item Automatic sleep staging: Given the sleep signal sequence $\mathbf{X}_k=[\mathbf{x}_{k,1},\mathbf{x}_{k,2},\cdots,\mathbf{x}_{k,T}]$, the goal is to learn a function $f: \mathbb{R}^{T \times N} \rightarrow \mathcal{Y}^T$ that maps $X_k$ to the sleep stage sequence $\mathbf{y}_k=\left [y_{k,1},y_{k,2},\cdots,y_{k,T}\right ]$
\end{itemize}

\begin{table}[t]
\centering
\caption{Notation definition}
\resizebox{0.6\textwidth}{!}{\begin{tabular}{ll}
\hline
Notation                       & Description           \\ \hline
$D^S \equiv \{ \left(\textbf{x}_{k}, y_k, d_k\right) \}_{k=1}^{|D^S|}$  & Labeled dataset ($1^{st}$ training stage)  \\ 
$D^U \equiv \{ \left(\textbf{x}_{k}, d_k\right) \}_{k=1}^{|D^U|}$  & Unlabeled dataset ($1^{st}$ training stage)  \\
$\textbf{x}_k$ & $k^{th}$ sleep segment \\
$y_{k}$& sleep stage for $\textbf{x}_{k}$       \\ 
$d_{k}$ & subject id for $\textbf{x}_{k}$       \\ 
$\widehat{\textbf{x}}_{k}$                      & $k^{th}$ augmented sleep segment            \\
$\widehat{y}_{k}$                       & sleep stage for $\widehat{\textbf{x}}_{k}$       \\
$\widehat{d}_{k}$                       & subject id for $\widehat{\textbf{x}}_{k}$ \\
$\textbf{z}^d_k, \textbf{z}^y_k$                       & latent representations for $\widehat{\textbf{x}}_{k}$ \\ 
$Q_{\phi_d}, Q_{\phi_y}$                       & encoder networks \\ 
$P_{\theta}$                       & decoder network \\
$P_{\theta_d}, P_{\theta_y}$                       & prior networks \\
$H_{\omega_d}, H_{\omega_y}$                       & classifiers \\
$\alpha_d, \alpha_y, \beta, \gamma_d, \gamma_y $                       & weights to control losses ($1^{st}$ training stage) \\

\hline
$D^S \equiv \{ \left(\textbf{X}_{k}, \textbf{y}_k\right) \}_{k=1}^{|D^S|}$  & Labeled dataset ($2^{nd}$ training stage)  \\ 
$\textbf{X}_k$  & $k^{th}$ sleep signal sequence  \\ 
$\textbf{x}_{k,i}$  & $i^{th}$ sleep segment in $\textbf{X}_k$       \\ 
$\textbf{y}_k$                       & sleep stage sequence for $\textbf{X}_k$                      \\ 
$y_{k,i}$     & sleep stage for $\textbf{x}_{k,i}$     \\ 
$\textbf{Z}_k$                       & representation sequence for $\widehat{\textbf{X}}_{k}$ \\
$\widehat{\textbf{Z}}_k$                       &  contextualized representation sequence for $\widehat{\textbf{X}}_{k}$ \\
$\textbf{y}_k^{*}$                       & latent predicted stage sequence for $\widehat{\textbf{X}}_{k}$ \\
$S_{\varphi}$ & potential function in CRF \\
$\mathcal{U}$ & uncertainty measure \\

\hline
\end{tabular}}
\label{notation}
\end{table}

\subsection{Proposed model}

The proposed DREAM is composed of a feature representation network that learns sleep-related and subject-invariant representations from the input sleep segment and a classification network that captures the interactions and correlations between these latent representations in the sequential context to predict the best corresponding sleep stage sequence. DREAM is optimized in two stages. First, the feature representation network is trained for each segment in both labeled and unlabeled datasets. Then, given a sequence of sleep segments, the trained feature network with fixed weights is employed to generate a sequence of vector representations, which is then fed into the classification network to train the classification network by predicting the sequence of sleep stages. The overall architecture of DREAM is shown in Figure \ref{fig:2}.

\subsubsection{Feature representation network}
We leverage the VAE framework~\cite{kingma2013auto} to learn sleep-related and subject-invariant representations for sleep segments. The model relies on an assumption about the sleep segment generation process in which the sleep signal $\textbf{x}$ is generated from two latent vectors $\textbf{z}_d$ and $\textbf{z}_y$ that capture only subject- and sleep-stage-related information, respectively. Then, under this assumption, if $\textbf{x}$ is completely disentangled into $\textbf{z}_d$ and $\textbf{z}_y$, $\textbf{z}_y$ becomes a sleep-related but subject-invariant representation. The training process is as follows. Given the labeled dataset $D^S \equiv \{ \left(\textbf{x}_{k}, y_k, d_k\right) \}_{k=1}^{|D^S|}$ where $d_k$ is subject id, we first generate two augmented examples $\widehat{\textbf{x}}_{2k}$ and $\widehat{\textbf{x}}_{2k+1}$ for each $\textbf{x}_k$ using data augmentation methods following~\cite{jiang2021self}: (i) randomly split the segment into chunks then randomly permute them, and (ii) randomly crop the segment and resize it to its original size using linear interpolation. The rationale behind data augmentation is to leverage the additional information provided by the augmented samples. These samples are generated with the intention of enhancing the model's ability to capture underlying patterns and features in the data. This process generates the augmented labeled dataset $\widehat{D}^S \equiv \left\{ \left(\widehat{\textbf{x}}_{k}, \widehat{y}_k, \widehat{d}_k\right) \right\}_{k=1}^{2|D^S|}$ where $\widehat{y}_{2k} = \widehat{y}_{2k+1} = y_k$ and $\widehat{d}_{2k} = \widehat{d}_{2k+1} = d_k$. Then, we maximize the ELBO with two latent variables $\textbf{z}_k^d$ and $\textbf{z}_k^y$  as a surrogate function for augmented data log-likelihood as follows.

\begin{align*}
    \mathcal{L}_{VAE} = &  \mathbb{E}_{Q_{\phi_d}\left(\textbf{z}^d_k|\widehat{\textbf{x}}_k\right), Q_{\phi_y}\left(\textbf{z}^y_k|\widehat{\textbf{x}}_k\right) } \left[ P_{\theta}\left(\widehat{\textbf{x}}_k|\textbf{z}^d_k, \textbf{z}^y_k\right) \right] \\
    & -\beta \KL \left(Q_{\phi_d}\left(\textbf{z}^d_k|\widehat{\textbf{x}}_k\right) \parallel P_{\theta_d}\left(\textbf{z}^d_k|\widehat{d}_k\right)\right) \\
    & -\beta \KL \left(Q_{\phi_y}\left(\textbf{z}^y_k|\widehat{\textbf{x}}_k\right) \parallel P_{\theta_y}\left(\textbf{z}^y_k|\widehat{y}_k\right)\right) 
\end{align*}

$Q_{\phi_d}$ and $Q_{\phi_y}$ are two encoder networks based on ResNet-50~\cite{he2016deep} that map input $\widehat{\mathbf{x}}_k$ into latent representations $\textbf{z}_k^d$ and $\textbf{z}_k^y$, respectively. $P_{\theta}$ is a decoder composed of 2-layer feed-forward layers followed by 3 transposed convolutional layers. $P_{\theta}$ reconstructs the input $\widehat{\mathbf{x}}_k$ from its latent representations $\textbf{z}_k^d$ and $\textbf{z}_k^y$. $P_{\theta_d}$ and $P_{\theta_y}$ are two prior networks composed of 3-layer feed-forward networks, for $\textbf{z}_k^d$ and $\textbf{z}_k^y$, respectively. $\KL(\cdot)$ denotes the Kullback–Leibler divergence (KL-divergence) loss between two distributions, and $\beta$ is a weight that controls KL-divergence constraints. Motivated by $\beta$-VAE model~\cite{higgins2016beta}, a larger value of $\beta$ forces each dimension of $\textbf{z}_k^d$ and $\textbf{z}_k^y$ to capture one of the conditionally independent factors in x. We also set two classifiers $H_{\omega_d}$ and $H_{\omega_y}$, 1-layer feed-forward networks, that predict $\widehat{d}_k$ with $\textbf{z}_k^d$ and $\widehat{y}_k$ with $\textbf{z}_k^y$, respectively, to further force the disentangled representation $\textbf{z}_k^d$ (resp. $\textbf{z}_k^y$) only to capture information about $\widehat{d}_k$ (resp. $\widehat{y}_k$). The classification losses are optimized as follows.

\begin{align*}
    &\mathcal{L}_{VAE}^{d} = - \mathbb{E}_{Q_{\phi_d}\left(\textbf{z}^d_k|\widehat{\textbf{x}}_k\right)}\left[ \log H_{\omega_d}\left(\widehat{d}_k|\textbf{z}^d_k\right) \right] \\
    &\mathcal{L}_{VAE}^{y} = - \mathbb{E}_{Q_{\phi_y}\left(\textbf{z}^y_k|\widehat{\textbf{x}}_k\right)}\left[ \log H_{\omega_y}\left(\widehat{y}_k|\textbf{z}^y_k\right) \right]
\end{align*}

To strengthen the robustness of the latent representations, self-supervised contrastive learning~\cite{chen2020simple} for $\textbf{z}^d_k$ and supervised contrastive learning~\cite{khosla2020supervised} for $\textbf{z}^y_k$ are applied. Contrastive learning aims to learn an embedding space where similar representations are close and dissimilar representations are far away. For subject-related representation $\textbf{z}^d_k$, its similar representations are the ones generated from its augmented examples. For sleep-related representation $\textbf{z}^y_k$, its similar ones are representations from the segments with the same stage. Given the set ${\textbf{z}_k^d}_{k=1}^{2|D^S|}$, the self-supervised contrastive objective function for a similar pair $\{\textbf{z}^d_k,\textbf{z}^d_i \}$ is expressed as 

\begin{equation*}
    \mathcal{L}_{CL}^d = -\log \frac{\exp\left(\simm\left(\psi_d\left(\textbf{z}^d_k \right),\psi_d\left(\textbf{z}^d_i \right)\right)/ \rho\right)}{\sum_{j \in A(k)}\exp\left(\simm\left(\psi_d\left(\textbf{z}^d_k \right),\psi_d\left(\textbf{z}^d_j \right)\right)/ \rho\right)}
\end{equation*}

where $A(k)\equiv\left\{1,2,\cdots,2\left|\mathcal{D}^S\right| \right\} \setminus \left\{k\right\}$ is a function that maps $\textbf{z}_{k}^{d}$ to the embedding space, $\simm(\textbf{u},\textbf{v})$ is the cosine similarity between two vectors $\textbf{u}$, $\textbf{v}$, and $\rho$ is a temperature parameter. For $\textbf{z}_{k}^{y}$, the supervised contrastive objective function is expressed as follows

\begin{equation*}
    \mathcal{L}_{SCL}^{y} = \frac{1}{|P(k)|} \times 
    \sum_{p \in P(k)} -\log \frac{\exp\left(\simm\left(\psi_y\left(\textbf{z}^y_k \right),\psi_y\left(\textbf{z}^y_p \right)\right)/ \rho\right)}{\sum_{j \in A(k)}\exp\left(\simm\left(\psi_y\left(\textbf{z}^y_k \right),\psi_y\left(\textbf{z}^y_j \right)\right)/ \rho\right)}
\end{equation*}
where $P(k) \equiv \{p \in A(k) : y_{p} = y_{k} \}$ is the set of indices of all positives in the data batch distinct from k and $|P(k)|$ is its cardinality. 

In sum, for the labeled dataset $D^S$, we optimize the following objective function.

\begin{equation}\label{eq:1}
    \mathcal{L}_{feature}^{S} = - \mathcal{L}_{VAE} + \alpha_d \mathcal{L}_{VAE}^{d} + \alpha_y \mathcal{L}_{VAE}^{y} + \gamma_d \mathcal{L}_{CL}^{d} + \gamma_y \mathcal{L}_{SCL}^{y}
\end{equation}

$\alpha_d,\alpha_y,\gamma_d,\gamma_y$ are hyper-parameters that control the relative importance of $\mathcal{L}_{VAE}^d,\mathcal{L}_{VAE}^y,\mathcal{L}_{CL}^d,\mathcal{L}_{SCL}^y$, respectively, compared to the $\mathcal{L}_{VAE}$. For the unlabeled dataset $D^{U} \equiv \{(x_k,d_k )\}_{k=1}^{|D^U|}$, we also generate an augmented unlabeled dataset $\hat{D}^U$with the augmentation methods mentioned previously, then optimize an objective function as follows.

\begin{equation}\label{eq:2}
    \mathcal{L}_{feature}^{U} = \gamma_d \mathcal{L}_{CL}^{d} 
\end{equation}

After training the VAE-based model with $\mathcal{L}_{feature}^S$ and $\mathcal{L}_{feature}^U$ on both labeled and unlabeled datasets, we use the trained encoder $Q_{\phi_y}$ with fixed weights as the feature representation network that extracts sleep-related and subject-invariant representations for input signals to train the sleep stage classification network.

\subsubsection{Classification network}
The classification model is composed of the Transformer and CRF. The subject-invariant and sleep-related representations from the trained $Q_{\phi_y}$ of the feature network, are input into the Transformer model to capture interactions between sequential sleep segments. Subsequently, the output of the Transformer is fed into the CRF model to find the best sequence of sleep stages, rather than labeling sleep segments independently. Specifically, given the dataset $D^S \equiv \{ \left(\textbf{X}_k, \textbf{y}_k \right)\}_{k=1}^{|D^S|}$, where $\mathbf{X}_k = \left [\mathbf{x}_{k,1}, \mathbf{x}_{k,2}, \cdots, \mathbf{x}_{k,T} \right ]$ is the sleep signal sequence, $\mathbf{y}_k=[y_{k,1},y_{k,2},\cdots,y_{k,T}]$ is the corresponding sleep stage sequence, and $T$ is the number of segments in $\mathbf{X}_k$ and $\mathbf{y}_k$, $\mathbf{X}_k$ is transformed to a sequence of latent representations, $\mathbf{Z}_k$, using the trained $Q_{\phi_y}$.

\begin{align*}
    \textbf{Z}_k &= [\textbf{z}_{k,1}, \textbf{z}_{k,2}, \cdots, \textbf{z}_{k,T}] \\
    &= \left[Q_{\phi_y}(\textbf{x}_{k,1}), Q_{\phi_y}(\textbf{x}_{k,2}), \cdots, Q_{\phi_y}(\textbf{x}_{k,T})\right]
\end{align*}

\begin{figure}[t]
    \centering
    \includegraphics[width=1\linewidth]{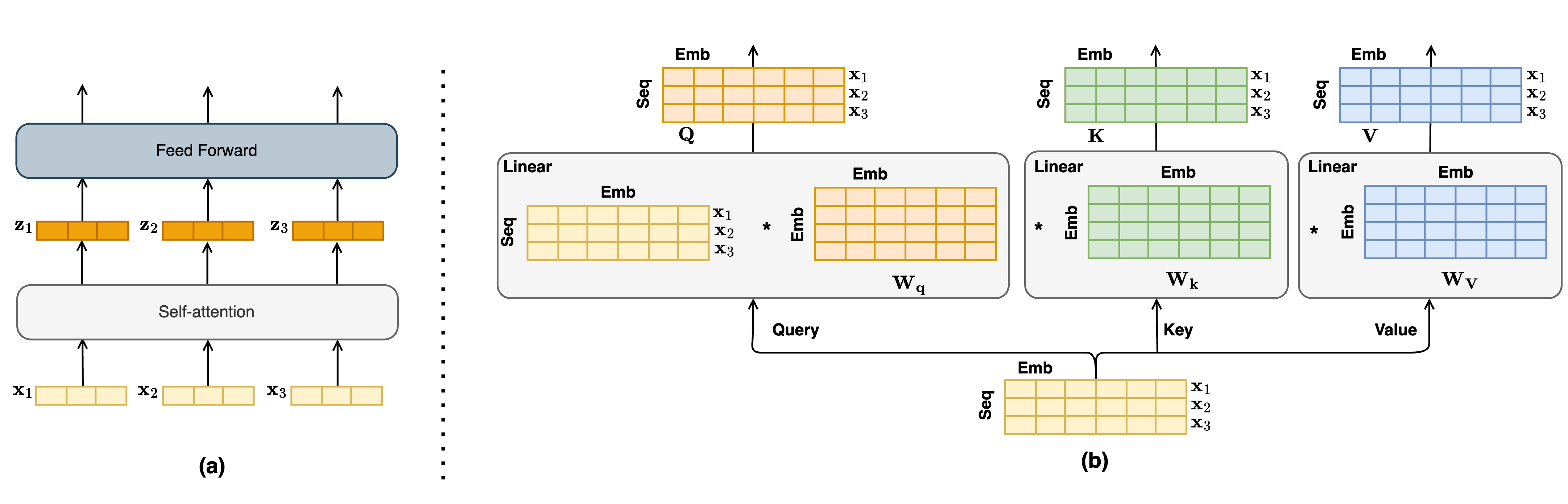}
    \caption{Visualization of (a) Transformer encoder and (b) Self-attention.}
    \label{fig:3}
\end{figure}

Then, to learn the contextualized representations $\widehat{\mathbf{Z}}_{k}$ for $\mathbf{Z}_k$, $\mathbf{Z}_k$ is fed into the 4-layer encoder network of the Transformer model with 8 attention heads for each layer as follows.

\begin{equation*}
    \widehat{\textbf{Z}}_k = \left[\widehat{\textbf{z}}_{k,1}, \widehat{\textbf{z}}_{k,2}, \cdots, \widehat{\textbf{z}}_{k,T}\right] = \transformer\left(\textbf{Z}_k\right)
\end{equation*}

Explicitly, each attention head in $\transformer$ extracts the contextualized representation sequence $\mathbf{Z}$ from the input sequence $\mathbf{X}$ as follow:

\begin{equation*}
    \textbf{Z} = \softmax \left ( \frac{\textbf{Q} \textbf{K}^T}{\sqrt{d_{k}}} \right) \textbf{V}
\end{equation*}
where $\mathbf{Q},\mathbf{K},\mathbf{V} \in R^{T \times d_{k}}$ are the query, key, value matrices calculated from $\mathbf{X}$, respectively, and $d_{k}$ denotes the dimension of the embedding space. The output of each $\transformer$ layer is concatenated from the outputs of attention heads in that layer. Given the contextualized representation $\widehat{\mathbf{Z}}_{k}$ from $\transformer$, the CRF model is then used to learn the best overall stage sequence, instead of finding a sleep stage for each sleep segment independently. Specifically, the CRF considers the correlations between the sleep stages of neighbors and jointly decodes the best chain of the stages for $\mathbf{X}_{k}$. It is helpful for accurate automatic sleep staging in the sequential context. For example, the N3 stage rarely follows the W or REM stage. If the model independently predicts sequential sleep stages, this constraint is difficult to be captured automatically and can further be violated. Formally, the CRF model computes the probability of the output sleep stage sequence $\textbf{y}_k$ as follows:

\begin{equation*}
    P_{\varphi}\left(\textbf{y}_k|\widehat{\textbf{Z}}_k\right) = \frac{\prod_{i=1}^{T}\mathcal{S}_{\varphi}\left(y_{k,i-1}, y_{k,i}, \widehat{\textbf{Z}}_k\right) }{\sum_{{\textbf{y}}' \in \mathcal{Y}^{T}} \prod_{i=1}^{T}\mathcal{S}_{\varphi}\left({y}'_{i-1}, {y}'_{i}, \widehat{\textbf{Z}}_k\right)}
\end{equation*}
where $S_{\varphi}\left (y_{k,i-1},y_{k,i},\widehat{\mathbf{Z}}_{k}\right )$ is the potential function constructed from a 1-layer feed-forward network that considers the correlation between $y_{k,i-1}$ and $y_{k,i}$ in given $\widehat{\mathbf{Z}}_k$. During the training phase, the negative conditional log-likelihood is minimized as follows.

\begin{equation}\label{eq:3}
    \mathcal{L}_{label} =  - \log P_{\varphi}\left(\textbf{y}_k|\widehat{\textbf{Z}}_k\right)
\end{equation}

In the inference phase, given $\widehat{\mathbf{Z}}_k$, the best sleep stage sequence with the highest conditional probability is selected by the Viterbi algorithm~\cite{forney1973viterbi}.

\begin{equation*}
    \textbf{y}_k^{*} = \underset{{\textbf{y}}' \in \mathcal{Y}^{T}}{\arg \max} P_{\varphi}\left({\textbf{y}}'|\widehat{\textbf{Z}}_k\right)
\end{equation*}

\begin{figure}[t]
    \centering
    \includegraphics[width=.65\linewidth]{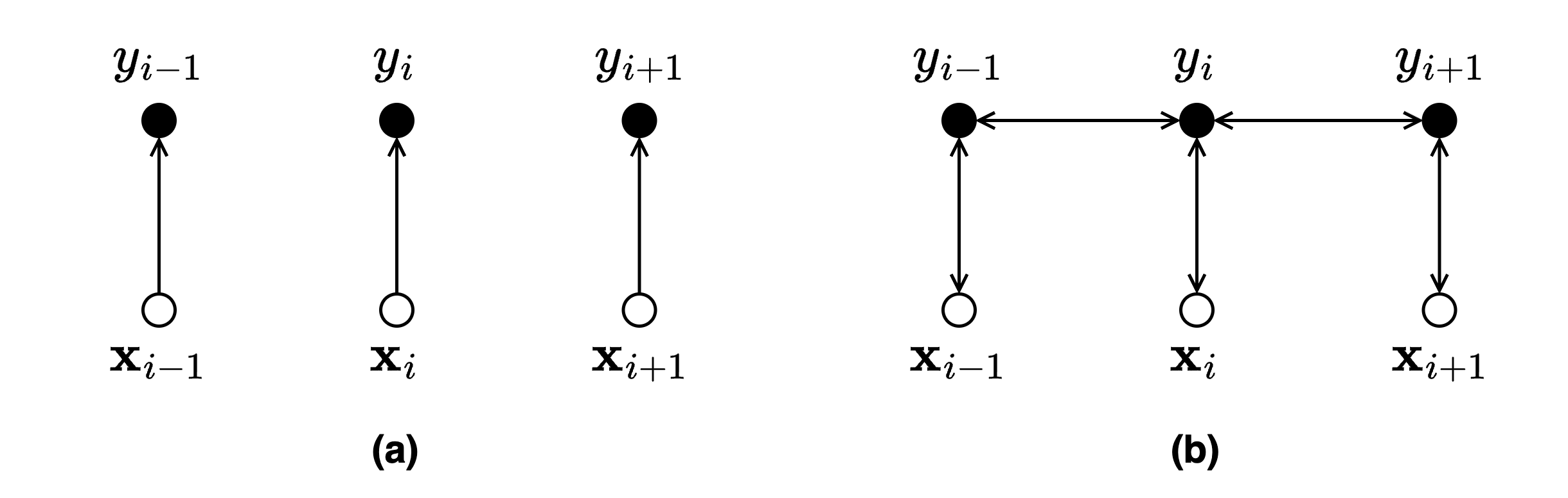}
    \caption{Visualization of predictions, showing (a) without CRF and (b) with CRF.}
    \label{fig:4}
\end{figure}

\begin{algorithm2e}[t]
\small
\caption{Two-stage training process for DREAM}\label{alg:fatdm}
\SetAlgoLined
\KwIn{Labeled dataset $\mathcal{D}^{S}$, unlabeled dataset $\mathcal{D}^{U}$}
\KwOut{Feature representation network $Q_{\theta_y}$, sleep stage classification network $f$}
\SetKwFunction{procone}{Feature\_Learning}
\SetKwFunction{proctwo}{Classification\_Learning}
\SetKwFunction{size}{count\_atoms}
\SetKwFunction{bond}{get\_brics\_bonds}
\SetKwFunction{brk}{break\_molecule}
\SetKwFunction{len}{length}
\SetKw{and}{and}
\SetKw{or}{or}
\SetKw{not}{not}
\SetKw{true}{true}
\SetKw{false}{false}
\SetKwProg{myproc}{Procedure}{}{}

\myproc{\procone{$\mathcal{D}^{S}$, $\mathcal{D}^{U}$}}{
  \For{epoch $=1$ to E}{
    \For{batch $\textbf{b} = \{ \left(\textbf{x}_{k}, d_k\right) \}_{k=1}^{|\textbf{b}|}$ in $\mathcal{D}^{U}$}{
      Generate augmented batch $\widehat{\textbf{b}}$ by data augmentation\;
      Optimize $\mathcal{L}_{feature}^{U}$ in Eq. (\ref{eq:2})\;
    }
    \For{batch $\textbf{b} = \{ \left(\textbf{x}_{k}, y_k, d_k\right) \}_{k=1}^{|\textbf{b}|}$ in $\mathcal{D}^{S}$}{
      Generate augmented batch $\widehat{\textbf{b}}$ by data augmentation\;
      Optimize $\mathcal{L}_{feature}^{S}$ in Eq. (\ref{eq:1})\;
    }
  }
}

\myproc{\proctwo{$\mathcal{D}^{S}$}}{
  \For{epoch $=1$ to E}{
    \For{batch $\textbf{b} = \{ \left(\textbf{X}_{k}, \textbf{y}_k\right) \}_{k=1}^{|\textbf{b}|}$ in $\mathcal{D}^{S}$}{
      Optimize $\mathcal{L}_{label}^{S}$ in Eq. (\ref{eq:3})\;
    }
  }
}
\end{algorithm2e}

Figure \ref{fig:4} shows a visual comparison of predictions, showing (a) without the CRF and (b) with the CRF. The difference between these two visualizations highlights the impact of CRF in improving the prediction accuracy, particularly in capturing sequential patterns within the data.

\subsubsection{Optimization}
The proposed DREAM is optimized by a 2-stage training process. First, the feature representation network $Q_{\phi_y}$ is trained by minimizing $L_{feature}^S$ on the labeled dataset and $\mathcal{L}_{feature}^U$ on the unlabeled dataset, respectively. Specifically, for each epoch, we train the model on the labeled data first, followed by the unlabeled data. Second, using the trained network $Q_{\phi_{y}}$ with fixed weights, the classification network $f$ is trained such that $\mathcal{L}_{label}$ is minimized on the labeled dataset. The details of this 2-stage optimization process are shown in Algorithm \ref{alg:fatdm}.

\subsubsection{Uncertainty quantification}
Uncertainty quantification is crucial for the deployment of machine learning models in clinical environments. Thus, we propose a new mechanism that measures the uncertainty of predicted sequence generated from CRF model. In particular, the uncertainty score of the prediction $y^{*}_{k,i}$ for the $i^{th}$ segment $\textbf{x}_{k,i}$ in the signal sequence $\textbf{X}_{k}$ is calculated as the entropy of the conditional distribution $P_{\varphi}\left(y^{*}_{k,i} | y^{*}_{k,i-1}, \widehat{\textbf{Z}}_k\right)$. For example, if $y^{*}_{k,i-1} = c$ with $c \in \mathcal{Y}$, then the uncertainty score of $y^{*}_{k,i}$ is computed as

\begin{equation*}
    \mathcal{U}\left(y^{*}_{k,i}\right) = \sum_{{c}' \in \mathcal{Y}} \left (  P_{\varphi}\left(y^{*}_{k,i} = {c}' | y^{*}_{k,i-1} = c, \widehat{\textbf{Z}}_k\right) \right.  \ast \log \left. \left( P_{\varphi}\left(y^{*}_{k,i} = {c}' | y^{*}_{k,i-1} = c, \widehat{\textbf{Z}}_k\right) \right) \right)
\end{equation*}
where
\begin{equation*}
    P_{\varphi}\left(y^{*}_{k,i} = {c}' | y^{*}_{k,i-1} = c, \widehat{\textbf{Z}}_k\right) = \frac{\mathcal{S}_{\varphi}\left(y_{k,i-1}=c, y_{k,i}={c}', \widehat{\textbf{Z}}_k\right)}{\sum_{{c}'' \in \mathcal{Y}} \mathcal{S}_{\varphi}\left(y_{k,i-1} = c, y_{k,i} = {c}'', \widehat{\textbf{Z}}_k\right)}
\end{equation*}

\section{Results}
This section reports the results from 5 forms of evaluation to demonstrate the effectiveness of the proposed method: (1) a quantitative experiment on sleep stage prediction tasks to compare our model with the state-of-the-art models, (2) a case study, (3) the usage of unlabeled data, (4) uncertainty quantification, and (5) the ablation study.

\subsection{Datasets}
Experiments are conducted on three real-world datasets for sleep studies: SleepEDF-20, SleepEDF-78 (an expansion of SleepEDF-20)~\cite{kemp2000analysis,goldberger2000physiobank},  and SHHS~\cite{quan1997sleep,zhang2018national}. SleepEDF datasets contain whole-night PSG with healthy subjects aged 25 to 101 to study the effect of age on sleep. The PSG consists of several signals including EEG, EOG, and EMG channels, and an event marker. The signals are divided into 30-second segments, and then each segment is labeled as one of eight classes {W, REM, N1, N2, N3, N4, MOVEMENT, UNKNOWN}. According to the previous studies~\cite{mousavi2019sleepeegnet, eldele2021attention, sokolovsky2019deep, phan2018joint}, to preprocess the SleepEDF datasets, we first extracted 30-second interval segments at a sampling rate of 100 Hz from single-channel EEG signals (Fpz-Cz), known to have the most information about sleep stages. Then, we excluded all segments with UNKNOWN and MOVEMENT stages and merged N4 into N3. SHHS is a multi-center cohort study to investigate the cardiovascular and other consequences of sleep-disordered breathing. Following the previous works~\cite{fonseca2016cardiorespiratory, eldele2021attention}, out of 6,441 subjects, only 329 subjects considered to have regular sleep were selected for our experiments. To construct the dataset used in the experiments, we extracted sleep segments at a sampling rate of 125 Hz from single-channel EEG signals (C4-A1). Table \ref{tb:1} shows the statistics of all datasets.

\subsection{Baselines}
To evaluate the performance of the proposed DREAM for automatic sleep staging, we compare DREAM with the following baselines, including 1 traditional machine learning model and 5 deep learning models. These models are detailed below.
\begin{itemize}
    \item Gradient Boosting Machine (GBM)~\cite{friedman2001greedy}. A classic ensemble model that performs prediction by averaging the predictions from many decision tree classifiers.
    \item ResNet+GRU. A neural network-based model composed of ResNet for feature representation and gated recurrent unit (GRU)~\cite{cho:2014}, which is a variant of RNNs, for classification.
    \item ResNet+LSTM. A neural network-based model composed of ResNet for feature representation and long short-term memory (LSTM)~\cite{graves2012long} for classification.
    \item DeepSleepNet~\cite{supratak2017deepsleepnet}. A neural network-based model consisting of CNN layers for representation learning and LSTM layers for classification. 
    \item U-time~\cite{perslev2019u}. A neural network model consisting only of CNN layers with skip connections.
    \item AttnSleep~\cite{eldele2021attention}. A neural network-based model consisting of CNN layers with various kernel sizes and multi-head attention.
\end{itemize}

\begin{table*}[t]
\centering
\caption{Statistics of sleep staging datasets.}
\label{tb:1}
\resizebox{1\textwidth}{!}{
\begin{tabular}{lc|ccc|ccccccc}
\hline
\multicolumn{1}{c}{\multirow{2}{*}{Datasets}} & \multicolumn{1}{c|}{\multirow{2}{*}{\# Subjects}} &  \multicolumn{3}{c|}{Statistics of EEG values}         & \multicolumn{6}{c}{Statistics of sleep stages}               \\ \cline{3-11} 
\multicolumn{1}{c}{} & & Avg. & Std.  & Range & W  & N1   & N2   & N3  & R & Total    \\ \hline
SleepEDF-20 & 20 & -1.2e-07 & 0.002 & (-0.021, 0.021)  & 8,285 (20\%) & 2,804 (7\%) & 17,799 (42\%) & 5,703 (13\%) & 7,717 (18\%)     & 42,308  \\ \hline
SleepEDF-78 & 78 & 1.2e-07  &   0.002 & (-0.025, 0.021) & 65,951 (34\%) & 21,522 (11\%) & 69,132 (35\%)  & 13,039 (7\%)        & 25,835 (13\%) & 195,479 \\ \hline
SHHS & 329& -6.6e-07  & 0.004 & (-0.016, 0.016) & \multicolumn{1}{l}{46,319 (14\%)} & \multicolumn{1}{l}{10,304 (3\%)} & \multicolumn{1}{l}{142,125 (44\%)} & \multicolumn{1}{l}{60,153 (19\%)} & \multicolumn{1}{l}{65,953 (20\%)} & 324,854 \\ \hline
\end{tabular}}
\end{table*}

\begin{table*}[t]
\centering
\caption{Comparison of prediction performance measured by accuracy, MF1, and $k$ scores on three sleep staging datasets. The average scores and the standard deviation between folds are reported.}
\label{tb:2}
\begin{tabular}{llccc}
\hline
\multirow{2}{*}{} & \multirow{2}{*}{Method} & \multicolumn{2}{c}{Overall Metrics} \\ \cline{3-5} 
 &  & Accuracy & MF1 & $k$ \\ \hline
SleepEDF-20 & DREAM & \textbf{83.91 $\pm$ 5.62} & \textbf{75.72 $\pm$ 6.06} & \textbf{0.77 $\pm$ 0.08} \\
 & AttnSleep & 81.31 $\pm$ 7.36 & 75.06 $\pm$ 7.47 & 0.74 $\pm$ 0.09 \\
 & DeepSleepNet & 80.75 $\pm$ 7.64 & 74.37 $\pm$ 7.45& 0.74 $\pm$ 0.10 \\
 & U-time & 78.81 $\pm$ 8.34 & 69.71 $\pm$ 8.14 & 0.70 $\pm$ 0.11 \\
 & ResNet+LSTM & 78.95 $\pm$ 7.55 & 66.92 $\pm$ 7.15 & 0.70 $\pm$ 0.10 \\
 & ResNet+GRU & 78.52 $\pm$ 7.46 & 65.03 $\pm$ 7.23 & 0.70 $\pm$ 0.10 \\
 & GBM & 57.21 $\pm$ 6.19 & 47.66 $\pm$ 5.76 & 0.40 $\pm$ 0.08 \\ \hline
SleepEDF-78 & DREAM & \textbf{81.94 $\pm$ 2.18} & \textbf{74.79 $\pm$ 2.56} & \textbf{0.75 $\pm$ 0.03} \\
 & AttnSleep & 79.09 $\pm$ 2.67 & 73.59 $\pm$ 2.75& 0.71 $\pm$ 0.04 \\
 & DeepSleepNet & 76.58 $\pm$ 4.70 & 72.65 $\pm$ 4.12 & 0.68 $\pm$ 0.06 \\
 & U-time & 74.84 $\pm$ 2.56 & 63.81 $\pm$ 3.34& 0.65 $\pm$ 0.04 \\
 & ResNet+LSTM & 78.14 $\pm$ 2.59 & 69.88 $\pm$ 2.87& 0.70 $\pm$ 0.04 \\
 & ResNet+GRU & 78.11 $\pm$ 2.28 & 69.54 $\pm$ 2.92& 0.69 $\pm$ 0.03 \\
 & GBM & 57.43 $\pm$ 6.73 & 43.60 $\pm$ 11.69& 0.38 $\pm$ 0.13 \\ \hline
SHHS & DREAM & \textbf{83.90 $\pm$ 0.78} & \textbf{75.72 $\pm$ 1.08} & \textbf{0.77 $\pm$ 0.01} \\
 & AttnSleep & 79.30 $\pm$ 2.25 & 71.25 $\pm$ 2.28& 0.71 $\pm$ 0.03 \\
 & DeepSleepNet & 76.81 $\pm$ 3.15 & 70.33 $\pm$ 2.36& 0.68 $\pm$ 0.04 \\
 & U-time & 79.22 $\pm$ 1.56 & 65.33 $\pm$ 3.04& 0.70 $\pm$ 0.02 \\
 & ResNet+LSTM & 80.56 $\pm$ 0.53 & 65.01 $\pm$ 0.61& 0.72 $\pm$ 0.01 \\
 & ResNet+GRU & 80.35 $\pm$ 0.93 & 64.72 $\pm$ 0.98& 0.72 $\pm$ 0.01 \\
 & GBM & 63.80 $\pm$ 0.33 & 51.00 $\pm$ 0.36& 0.48 $\pm$ 0.00 \\ \hline
\end{tabular}
\end{table*}

\begin{table*}[t]
\centering
\caption{Comparison of prediction performance as measured by F1 score per class. The average scores and the standard deviation between folds are reported.}
\label{tb:3}
\begin{tabular}{clccccc}
\hline
\multirow{2}{*}{} & \multirow{2}{*}{Method} & \multicolumn{5}{c}{F1-score per class} \\ \cline{3-7} 
 &  & W & N1 & N2 & N3 & R \\ \hline
SleepEDF-20 & DREAM & \textbf{87.94 $\pm$ 8.07} & 37.22 $\pm$ 13.77 & \textbf{86.92 $\pm$ 6.20} & 85.32 $\pm$ 8.36 & \textbf{81.18 $\pm$ 9.15} \\
 & AttnSleep & 85.28 $\pm$ 11.07 & \textbf{39.08 $\pm$ 13.31} & 86.69 $\pm$ 7.96 & \textbf{88.28 $\pm$ 4.51} & 75.95 $\pm$ 11.00 \\
 & DeepSleepNet & 85.38 $\pm$ 9.73 & 37.11 $\pm$ 12.69 & 85.83 $\pm$ 9.57 & 87.92 $\pm$ 5.58 & 75.63 $\pm$ 10.69 \\
 & U-time & 80.79 $\pm$ 10.83 & 28.58 $\pm$ 10.72 & 83.45 $\pm$ 10.56 & 84.06 $\pm$ 10.82 & 71.64 $\pm$ 16.73 \\
 & ResNet+LSTM & 82.73 $\pm$ 9.76 & 13.15 $\pm$ 12.34 & 84.40 $\pm$ 8.60 & 84.55 $\pm$ 7.25 & 69.75 $\pm$ 12.68 \\
 & ResNet+GRU & 81.37 $\pm$ 9.99 & 5.95 $\pm$ 8.93 & 84.53 $\pm$ 8.21 & 83.42 $\pm$ 8.47 & 69.87 $\pm$ 12.78 \\
 & GBM & 54.91 $\pm$ 11.00 & 12.34 $\pm$ 6.50 & 66.32 $\pm$ 7.69 & 67.66 $\pm$ 12.94 & 37.10 $\pm$ 9.07 \\ \hline
SleepEDF-78 & DREAM & \textbf{91.82 $\pm$ 2.04} & 41.77 $\pm$ 4.02 & \textbf{84.68 $\pm$ 2.81} & 78.38 $\pm$ 5.57 & \textbf{77.32 $\pm$ 6.18} \\
 & AttnSleep & 90.92 $\pm$ 2.44 & 43.66 $\pm$ 4.11 & 82.62 $\pm$ 4.04 & \textbf{78.71 $\pm$ 7.81} & 72.02 $\pm$ 5.97 \\
 & DeepSleepNet & 86.80 $\pm$ 5.49 & \textbf{46.49 $\pm$ 5.43} & 79.54 $\pm$ 6.79 & 77.17 $\pm$ 6.19 & 73.26 $\pm$ 8.49 \\
 & U-time & 89.19 $\pm$ 2.61 & 20.57 $\pm$ 3.42 & 80.27 $\pm$ 3.87 & 74.03 $\pm$ 7.88 & 55.00 $\pm$ 6.34 \\
 & ResNet+LSTM & 89.45 $\pm$ 2.55 & 34.30 $\pm$ 4.24 & 82.60 $\pm$ 3.03 & 76.87 $\pm$ 7.94 & 66.17 $\pm$ 6.39 \\
 & ResNet+GRU & 89.86 $\pm$ 2.46 & 33.30 $\pm$ 4.85 & 82.40 $\pm$ 3.02 & 75.72 $\pm$ 7.61 & 66.41 $\pm$ 6.50 \\
 & GBM & 68.89 $\pm$ 12.37 & 13.50 $\pm$ 7.11 & 64.76 $\pm$ 5.42 & 48.14 $\pm$ 25.13 & 22.69 $\pm$ 11.75 \\ \hline
SHHS & DREAM & \textbf{85.16 $\pm$ 1.68} & \textbf{39.19 $\pm$ 2.16} & \textbf{85.78 $\pm$ 0.68} & 82.59 $\pm$ 1.35 & \textbf{85.87 $\pm$ 1.43} \\
 & AttnSleep & 81.35 $\pm$ 2.15 & 30.00 $\pm$ 3.10 & 84.59 $\pm$ 0.54 & \textbf{82.84 $\pm$ 2.26} & 77.45 $\pm$ 4.55 \\
 & DeepSleepNet & 76.91 $\pm$ 2.93 & 33.73 $\pm$ 4.78 & 79.63 $\pm$ 3.75 & 79.22 $\pm$ 1.94 & 82.16 $\pm$ 3.44 \\
 & U-time & 76.24 $\pm$ 3.81 & 10.47 $\pm$ 12.90 & 83.24 $\pm$ 1.02 & 76.01 $\pm$ 1.22 & 80.67 $\pm$ 2.64 \\
 & ResNet+LSTM & 80.83 $\pm$ 1.32 & 0.00 $\pm$ 0.00 & 83.67 $\pm$ 0.25 & 81.59 $\pm$ 1.97 & 78.98 $\pm$ 0.87 \\
 & ResNet+GRU & 79.23 $\pm$ 2.12 & 0.00 $\pm$ 0.00 & 83.73 $\pm$ 0.48 & 82.39 $\pm$ 0.76 & 78.27 $\pm$ 2.53 \\
 & GBM & 58.68 $\pm$ 1.23 & 2.53 $\pm$ 0.24 & 69.48 $\pm$ 0.60 & 73.05 $\pm$ 1.19 & 51.27 $\pm$ 1.61 \\ \hline
\end{tabular}
\end{table*}

\subsection{Implementation Details}
The machine learning and neural network-based methods are implemented by Scikit-Learn and PyTorch~\cite{paszke:2019}, respectively. We use ADAM algorithm~\cite{kingma:2014} to optimize the prediction performances for neural network-based models. For the neural network-based methods, we set the hyper-parameters according to the authors' implementation. For the proposed method, the batch size and the initial learning rate are set to 64 and 0.001, respectively. Other hyper-parameters for the feature network loss are set to $\alpha_y=3500,\alpha_d=10500,\gamma_d=\gamma_y=20000,\beta=1$ for the SleepEDF datasets and set to $\alpha_y=1000,\alpha_d=3000,\gamma_d=\gamma_y=2000$ for the SHHS dataset. For these hyper-parameters, we find the optimal values for each dataset, which show the best performance on the validation set. EDF-20 and -78 datasets have the same optimal hyper-parameters, whereas SHHS has different values. This difference suggests that the choice of hyper-parameters depends on the inherent characteristics of the data.

\subsection{Evaluation Metric}
We perform k-fold cross-validation method for our experiments on all datasets. For the SleepEDF-20 dataset, we use 20-fold cross-validation where each fold includes only one subject. The validation set is also divided for model optimization, so the ratio of train, validation, and test is 15:4:1. In the case of the SleepEDF-78 dataset, we conduct 10-fold cross-validation with 7 or 8 subjects in each fold. The ratio of train, validation, and test is approximately 9:1:1. In the case of the SHHS dataset, 5-fold cross-validation is used with a ratio of 3:1:1. To evaluate model performance, we use three overall evaluation metrics: accuracy, macro-averaged F1 score (MF1), and Cohen’s Kappa ($\kappa$)~\cite{cohen1960coefficient}. In addition to the overall evaluations, the F1 score for each sleep stage is also reported.

\subsection{Results of Automatic Sleep Staging}
Table \ref{tb:2} shows the overall performance as measured by accuracy, MF1, and $\kappa$ scores for sleep staging. The average score and the standard deviation between folds are reported. DREAM outperforms other baselines in the overall evaluation metrics. Especially, DREAM achieves the accuracies of 83.91\%, 81.94\%, and 83.90\% on SleepEDF-20, SleepEDF-78, and SHHS datasets, respectively, which are 3.4\% better on average than the second-best model (AttnSleep). Table \ref{tb:3} shows the F1 score for each sleep stage. DREAM consistently achieves the best performances for the three stages, W, N2, and R, across all datasets. Even on other stages, DREAM also achieves high performance. For the N1 stage, DREAM performs first-, second-, and third-best on SHHS, SleepEDF-20, and SleepEDF-78 datasets, respectively. In the case of the N3 stage, DREAM performs second-best on the SleepEDF-78 dataset and the third-best on SleepEDF-20 and SHHS datasets. We also perform statistical t-testing between DREAM and the baselines and calculate $p$ values based on accuracy scores. For all the baselines, $p$ values are less than $.001 (P<.001)$, indicating significant differences in the prediction performances between DREAM and the baselines. The experiment results demonstrate the competitiveness and the advantages of DREAM using (i) the feature representation network that learns sleep-related and subject-invariant representation for each sleep segment and (ii) the sleep stage classification network that models the complex interactions between the latent representations and between the sleep stages in the sequential context.

\begin{table}[t]
\centering
\caption{Prediction performances of DREAM and baseline models for worst-case testing subjects determined by EMD.}
\label{tab:4}
\begin{tabular}{llccc}
\hline
Testing subjects  & Models   & Accuracy  & MF1   & $k$  \\ \hline
\multirow{4}{*}{Subject \#4} 
 & DREAM        & \textbf{79.31} & \textbf{66.96} & \textbf{0.72} \\
 & AttnSleep    & 69.15 & 63.64 & 0.61 \\
 & DeepSleepNet & 65.54 & 60.63 & 0.57 \\
 & U-time       & 71.98 & 60.98 & 0.62 \\ \hline
\multirow{4}{*}{Subject \#12}   
 & DREAM        & \textbf{69.42} & \textbf{62.69} & \textbf{0.55} \\
 & AttnSleep    & 62.31 & 56.13 & 0.49 \\
 & DeepSleepNet & 61.04 & 55.71 & 0.49 \\
 & U-time       & 60.24 & 55.76 & 0.47 \\ \hline
\end{tabular}
\end{table}

\subsection{Worst-case performances}
We further investigate the effectiveness of learning sleep-related and subject-invariant representations for automatic sleep staging. It is expected that by capturing sleep-related and subject-invariant representations, DREAM will generalize its prediction performances to the testing subjects, especially in case there are differences between training and testing data distributions. To verify this hypothesis, we compare DREAM with baseline methods under the worst-case scenario in which a testing data/subject is significantly different from training data/subjects on the SleepEDF-20 dataset. To determine the worst-case testing subjects, we measure the distances between them and subjects in the training datasets and select testing subjects with the largest distances. In particular, for each testing fold under the cross-validation setting, we measure the distance between the testing subject with each subject in the training set by the earth mover's distance (EMD)~\cite{ramdas2017wasserstein} with respect to EEG signal spaces and take the average over all training subjects as the distance score. This process results in two worst-case subjects, subjects 4 and 12, with corresponding EMD scores of $9.23×10^{-4}$ and $6.66×10^{-4}$ compared to an averaged EMD score of $3.32×10^{-4}$ for all testing subjects. The prediction performances of DREAM and other baseline models for these two subjects are shown in Table 4. According to the accuracy measure, DREAM improves the prediction performances by about 15\% and 12\% for subjects 4 and 12 compared to the best baseline, which are significantly higher than the average improvement of 4\% over all subjects. These results indicate that our model can learn the generalized decision function based on sleep-relevant and subject-invariant features resulting in good prediction performances for the new subjects, especially in case there are differences between testing and training subjects.

\begin{table*}[t]
\centering
\caption{Prediction performances of DREAM with and without using unlabeled EEG signal data.}
\label{tab:5}
\resizebox{1\textwidth}{!}{
\begin{tabular}{llcccccccc}
\hline
\multicolumn{1}{c}{\multirow{2}{*}{Datasets}} & \multicolumn{1}{c}{\multirow{2}{*}{Setting}} & \multicolumn{3}{c}{Overall Metrics}  & \multicolumn{5}{c}{F1-score per class}                           \\ \cline{3-10} 
\multicolumn{1}{c}{}                          & \multicolumn{1}{c}{}                        & \multicolumn{1}{c}{Accuracy} & \multicolumn{1}{c}{MF1}   & $k$    & \multicolumn{1}{c}{W}     & \multicolumn{1}{c}{N1}    & \multicolumn{1}{c}{N2}    & \multicolumn{1}{c}{N3}    & R     \\ \hline
SleepEDF-20         & Supervised Learning & \multicolumn{1}{c}{83.91}    & \multicolumn{1}{c}{75.72} & 0.77 & \multicolumn{1}{c}{87.94} & \multicolumn{1}{c}{37.22} & \multicolumn{1}{c}{86.92} & \multicolumn{1}{c}{85.32} & 81.18 \\ \hline
SleepEDF-20 + SleepEDF-78                       & Semi-supervised Learning                        & \multicolumn{1}{c}{84.53}    & \multicolumn{1}{c}{76.17} & 0.78 & \multicolumn{1}{c}{88.71} & \multicolumn{1}{c}{37.05} & \multicolumn{1}{c}{87.60} & \multicolumn{1}{c}{85.57} & 81.90 \\ \hline
\end{tabular}}
\end{table*}

\subsection{The Usage of Unlabeled Data}
In this experiment, we look into the effect of using unlabeled EEG signal data on the prediction performances of our model. We consider the SleepEDF-20 dataset as the labeled dataset and the remaining data in the SleepEDF-78 dataset as the unlabeled dataset and use both of these two datasets in the training of the feature representation network. Specifically, we first optimize $\mathcal{L}_feature^U$ on the unlabeled dataset and then optimize $\mathcal{L}_feature^S$ on the labeled dataset for each epoch. Finally, the labeled dataset is used to train the classification network. As shown in Table \ref{tb:5}, using the unlabeled data improves the average accuracy of DREAM for testing subjects from 83.91\% to 84.53\%. This result indicates the benefit of leveraging unlabeled EEG data which is easy to collect for automatic sleep staging. We also conduct $t$-tests to evaluate the statistical significance of incorporating unlabeled data to train the model in accuracy scores. Table 5 shows the $p$-value when compared to the supervised model. The finding indicates that the use of unlabeled data has a statistically significant positive effect on the performance of sleep stage classification at a significance level of 0.05.

\begin{figure}[t]
    \centering
    \includegraphics[width=1.\linewidth]{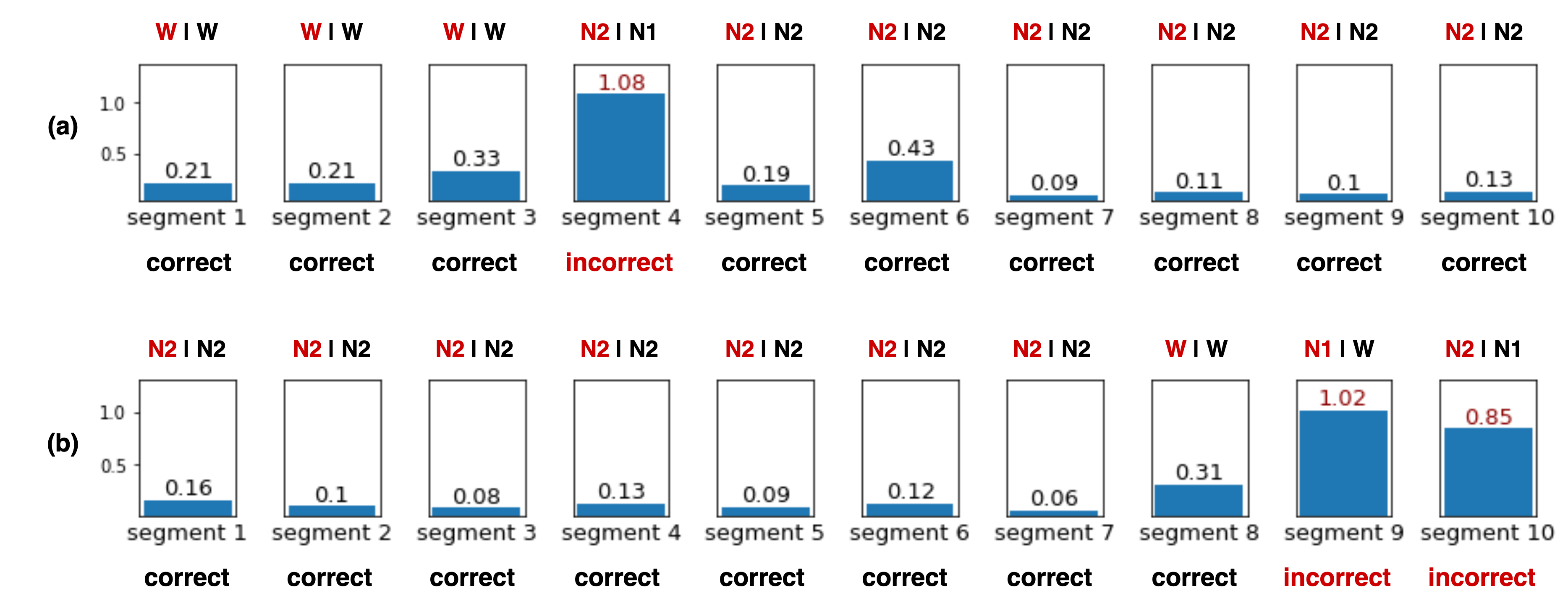}
    \caption{Case studies using uncertainty quantification to identify incorrect predictions. The figure shows uncertainty scores of sleep stage predictions for segments in the signal sequence. The prediction and ground-truth labels are marked with red and black colors, respectively. (a) the model makes an incorrect prediction at the 4th segment when the sleep cycle changes (W → N1 → N2). (b) the model makes incorrect predictions at the 9th and 10th segments when the sleep cycle changes (W → N1).}
    \label{fig:5}
\end{figure}

\subsection{Uncertainty Quantification}
Providing prediction only limits the use of machine learning models in clinical settings due to the difficulties of model debugging and human intervention. In this experiment, we investigate the capability of quantifying the prediction uncertainty of our model. Figure \ref{fig:5} shows the predictions and the corresponding uncertainty scores of two sleep signal sequences. In Figure \ref{fig:5}(a), the model makes the incorrect prediction at the 4th segment in the sequence when the sleep cycle moves from the W to N1 stages. In this segment, the uncertainty score is 1.08 which is significantly higher than the uncertainty scores for predicting other segments. In Figure \ref{fig:5}(b), the model makes incorrect predictions at the 9th and 10th when the sleep cycle moves from the W to N1 stages. Similar to Figure \ref{fig:4}(a), the uncertainty scores of the incorrect predictions are significantly higher than those of the correct predictions. These results show that the proposed uncertainty quantification method used in the CRF model can alert when sleep specialists need to look at the model predictions, thereby making it more reliable in the real scenario.

\subsection{Ablation Study}
We conduct an ablation study to examine the contribution of each component to the prediction performances of our model. In particular, starting from DREAM, each component is independently excluded from it to construct some model variants including DREAM w/o CRF (i.e., model uses feed-forward layer instead of CRF model to make predictions), DREAM w/o $\mathcal{L}_{VAE}, \mathcal{L}_{VAE}^d, \mathcal{L}_{VAE}^y$ (i.e., model without learning sleep-relevant and subject-invariant features), DREAM w/o $\mathcal{L}_{SCL}$ (i.e., model without supervised contrastive learning for feature representations), and DREAM w/o Aug (i.e., model without the data augmentation). As shown in Table \ref{tb:6}, the prediction performance is reduced when removing each component from DREAM. This result further demonstrates the effectiveness of learning robust feature representations by the VAE-based architecture and contrastive learning and modeling sleep dynamics in the sequential context at the prediction level by the CRF model. We also perform t-tests to assess the statistical significance of these variations in accuracy scores. Table \ref{tb:6} shows the $p$-values compared to DREAM. These results demonstrate that each variation has a statistically significant positive impact on the performance of the sleep stage classification at a significance level of 0.05. The consistent improvement across variations underscores the importance of carefully considering and optimizing each component of the model architecture.

\begin{table}[t]
\centering
\caption{Ablation study for DREAM on SleepEDF-20 dataset.}
\label{tb:6}
\begin{tabular}{lcccc}
\hline
Models        & Accuracy & $MF_1$ & $k$ & $p$\\ \hline
DREAM         & 83.91 $\pm$ 5.62  & 75.72 $\pm$ 6.06 & 0.77 $\pm$ 0.08 & \\ 
DREAM w/o CRF & 81.22 $\pm$ 6.97   & 67.55 $\pm$ 6.28 & 0.74 $\pm$ 0.09 & 0.001 \\ 
DREAM w/o $\mathcal{L}_{VAE}, \mathcal{L}_{VAE}^{d}, \mathcal{L}_{VAE}^{y}$ 
& 81.13 $\pm$ 7.24  & 71.43 $\pm$ 9.18  & 0.73 $\pm$ 0.11 & 0.000 \\ 
DREAM w/o $\mathcal{L}_{SCL}$ & 81.56 $\pm$ 7.39 &  72.86 $\pm$ 7.44 & 0.74 $\pm$ 0.10 & 0.006 \\ \hline
\end{tabular}
\end{table}

\begin{table}[]
\centering
\caption{Ablation study for the sleep stage classification network of DREAM on SleepEDF-20 dataset.}
\label{tb:7}
\begin{tabular}{lcccc}
\hline
Models & Accuracy & MF & $k$ & $p$   \\ \hline
DREAM                      & 83.91 $\pm$ 5.62 & 75.72 $\pm$ 6.06 & 0.77 $\pm$ 0.08 &       \\
FeatureNet w/ Logistic+CRF & 81.84 $\pm$ 6.86 & 71.18 $\pm$ 7.35 & 0.74 $\pm$ 0.09 & 0.003 \\
FeatureNet w/ Logistic     & 63.81 $\pm$ 9.83 & 43.13 $\pm$ 9.99 & 0.47 $\pm$ 0.13 & 0.000 \\ \hline
\end{tabular}
\end{table}

Additionally, we conduct experiments using logistic models with/without CRF as alternatives to our sleep stage classification network, which consists of the Transformer and CRF. The results are summarized in Table \ref{tb:7}. DREAM consistently outperforms these alternatives on all metrics. This demonstrates the significance and effectiveness of the sleep stage classification network in achieving the best results. While the Transformer can extract sequential dependencies in the input data, which is critical for sleep stage classification, the logistic model lacks the ability to capture sequential information. Therefore, our sleep stage classification network with the Transformer with the CRF is more suitable for accurate sleep stage classification, as it leverages the strengths of both methods to handle feature extraction and sequential dependencies in the data.

\subsection{Transferability}
Transferability is a crucial factor in assessing the model's adaptability in various data environments and its ability to successfully generalize. We conduct additional experiments to evaluate the model's transferability. However, it is challenging to directly assess transferability since the difference in sampling rates between the EDF and SHHS datasets leads to distinct model structures. To address this issue, we employ a two-step approach to evaluate transferability. First, the model is pre-trained on one dataset, and then it is fine-tuned using another dataset. Specifically, we fine-tune the sleep-stage-related encoder $Q_{\theta_y}$ (the number of nodes depends on the sampling rate) and the sleep stage classifier $H_{\omega_y}$ in the feature network, and the classification network. Table \ref{tb:8} shows the results, where "train" represents the dataset used for pre-training, and "test" indicates the dataset for fine-tuning and testing. As demonstrated in the table, the pre-trained model improves the performance when applied to different datasets.

\section{Discussion}
\subsection{Principal Results}
In this study, we present DREAM, a deep learning-based automatic sleep staging model, which learns subject-invariant and sleep-related representations for various sleep signal segments in both labeled and unlabeled datasets and models sleep dynamics by capturing complex interactions between the signal segments and between sleep stages in the sequential context. DREAM also provides an efficient mechanism to quantify the prediction uncertainty. Comprehensive experiments on real-world datasets demonstrate that DREAM not only outperforms state-of-the-art models, but also generalizes prediction performance for new subjects that differ significantly from training subjects. This suggests the trained DREAM can be used efficiently for other cohorts. Especially, uncertainty quantification further demonstrates the effectiveness of DREAM in real-world applications, as it gives sleep experts detailed information about predictions to help them make better decisions. Additionally, the ablation study evaluates the importance of various model components, highlighting the model's competitiveness and advantages in sleep stage classification. PSG data is relatively easy to collect, but annotating it is highly challenging. Our experimental findings that unlabeled data improve prediction performance provide many research opportunities to explore and pursue the benefits of unlabeled data.

\subsection{Challenges in Sleep Stage Classification}
The consistently low F1 scores observed for the N1 stage across all models, including the proposed one, raise important considerations regarding the classification of this particular sleep stage. Several factors contribute to the difficulty in accurately distinguishing the N1 stage. First and foremost, the limited availability of N1 stage data poses a significant challenge. Machine learning models heavily rely on a sufficient amount of diverse training data to generalize effectively. In many datasets, the N1 stage is underrepresented, leading to a lack of examples for the model to learn from. Furthermore, the intrinsic nature of the N1 stage may also contribute to its poor classification performance. The N1 stage often exhibits characteristics that are not well-distinguished from other sleep stages. These transitional states can be challenging to discern solely based on physiological signals, making it a complex classification task. Therefore, the difficulty of the N1 stage classification highlights the need for further research and innovation in this area. As the understanding of sleep physiology and data collection methods continue to evolve, addressing the challenges associated with the N1 stage classification will be crucial for improving the overall performance of sleep stage classification models. Additionally, incorporating more contextual information or multimodal data sources may enhance the model's ability to differentiate the N1 stage accurately.

\begin{table}[]
\centering
\caption{Transferability evaluation with fine-tuning}\label{tb:8}
\begin{tabular}{llccc}
\hline \multicolumn{2}{c}{Dataset} & \multicolumn{3}{c}{Overall Metrics} \\ \hline
Train        & Test         & Accuracy       & MF1      & $k$      \\ \hline
SHHS         & SHHS         & 83.90 $\pm$ 0.78          & 75.72 $\pm$ 1.08       &  0.77 $\pm$ 0.01     \\
SleepEDF-78  & SHHS         & 84.49 $\pm$ 0.32         & 76.11 $\pm$ 0.73     & 0.78 $\pm$ 0.00      \\
SleepEDF-78  & SleepEDF-78  & 81.94 $\pm$ 2.18             & 74.79 $\pm$ 2.56       & 0.75 $\pm$ 0.03      \\
SHHS         & SleepEDF-78  & 82.51 $\pm$ 2.03             & 75.27 $\pm$ 2.40       & 0.76 $\pm$ 0.02   \\   \hline
\end{tabular}
\end{table}

\section{Conclusions}
Automatic sleep staging has become a critical task in diagnosing and treating sleep disorders. In this paper, we propose a novel neural network-based framework named DREAM for automatic sleep staging. To address the challenges in sleep studies, DREAM learns sleep-related but subject-invariant representations under domain generalization and then effectively models the interactions between sequential sleep segments and between sleep stages. DREAM is optimized in two stages, where the feature network is first trained on both labeled and unlabeled datasets for representation learning, and then the classification network is optimized on the labeled data using the trained feature network. The experiments on real-world datasets demonstrate the advantages of DREAM. DREAM achieves higher performance compared to state-of-the-art baselines. In particular, the case study shows that our model can learn the generalized decision function resulting in generalized prediction performances for new subjects, especially in case there are differences between testing and training subjects. The experiments for the usage of unlabeled data demonstrate the benefit of leveraging unlabeled EEG data. More importantly, DREAM provides an efficient mechanism to quantify the prediction uncertainty, thereby making this model reliable and helping sleep specialists to make better decisions in real-world scenarios.

\section{Acknowledgments}
This work was funded in part by the National Science Foundation under award numbers IIS-2145625 and CBET-2037398.

\bibliographystyle{unsrt}  
\bibliography{references}

\end{document}